\documentclass[conference]{IEEEtran}
\IEEEoverridecommandlockouts
\usepackage{booktabs} 
\usepackage{graphicx}
\usepackage{setspace}
\usepackage{amsmath}
\usepackage{dsfont}
\usepackage{amsfonts}
\usepackage{lipsum}
\usepackage{multicol}
\usepackage{float}
\usepackage{bm}
\usepackage{color}
\usepackage[dvipsnames]{xcolor}
\usepackage{etoolbox}
\usepackage{soul}
\usepackage{amssymb}
\usepackage[linesnumbered,ruled,vlined]{algorithm2e}
\usepackage{mathtools,amssymb}
\usepackage{caption}
\usepackage{float}
\usepackage{amsthm}
\usepackage[caption = false,subrefformat=parens,labelformat=parens]{subfig}
\usepackage{adjustbox}
\usepackage{afterpage}
\usepackage{mathrsfs}
\usepackage{booktabs}
\usepackage{tabularx}
\usepackage{multirow}
\usepackage{array}
\usepackage{newtxtext,newtxmath}
\def\BibTeX{{\rm B\kern-.05em{\sc i\kern-.025em b}\kern-.08em
    T\kern-.1667em\lower.7ex\hbox{E}\kern-.125emX}}
\begin{document}

\title{A Unified Benchmark of Federated Learning with Kolmogorov–Arnold Networks for Medical Imaging

\author{
Youngjoon Lee\textsuperscript{\rm 1}, Jinu Gong\textsuperscript{\rm 2}, Joonhyuk Kang\textsuperscript{\rm 1} \\
\textsuperscript{\rm 1}School of Electrical Engineering, KAIST, South Korea\\
\textsuperscript{\rm 2}Department of Applied AI, Hansung University, South Korea\\
Emails: yjlee22@kaist.ac.kr, jinugong@hansung.kr, jkang@kaist.ac.kr
}

\thanks{This research was supported by the Institute of Information \& Communications Technology Planning \& Evaluation (IITP)-ITRC (Information Technology Research Center) grant funded by the Korea government (MSIT) (IITP-2025-RS-2020-II201787).\\
}
}

\maketitle

\begin{abstract}
Federated Learning (FL) enables model training across decentralized devices without sharing raw data, thereby preserving privacy in sensitive domains like healthcare. 
In this paper, we evaluate Kolmogorov–Arnold Networks (KAN) architectures against traditional MLP across six state-of-the-art FL algorithms on a blood cell classification dataset. 
Notably, our experiments demonstrate that KAN can effectively replace MLP in federated environments, achieving superior performance with simpler architectures. 
Furthermore, we analyze the impact of key hyperparameters—grid size and network architecture—on KAN performance under varying degrees of Non-IID data distribution. 
In addition, our ablation studies reveal that optimizing KAN width while maintaining minimal depth yields the best performance in federated settings. 
As a result, these findings establish KAN as a promising alternative for privacy-preserving medical imaging applications in distributed healthcare. 
To the best of our knowledge, this is the first comprehensive benchmark of KAN in FL settings for medical imaging task.
\end{abstract}


\section{Introduction}
Recent years have seen major progress in AI for healthcare, showing great potential across many medical uses \cite{bisio2025ai, bisio2023feet}. 
At the same time, growing privacy concerns have restricted access to medical data, leading to more interest in methods that protect privacy \cite{antunes2022federated}. 
Federated Learning (FL) is a new approach that allows training on spread-out datasets without sharing the actual data \cite{mcmahan2017communication, li2019convergence}. 
This has made FL popular in healthcare since it meets the strict privacy rules for medical data \cite{pfitzner2021federated, lee2024security}. 
However, FL performance hinges on the chosen model and its efficiency in distributed settings \cite{kairouz2021advances, lee2025revisit}. 
Therefore, studying effective model designs for FL remains important for advancing medical applications.

\begin{figure}[t]
\centering
\includegraphics[width=\columnwidth]{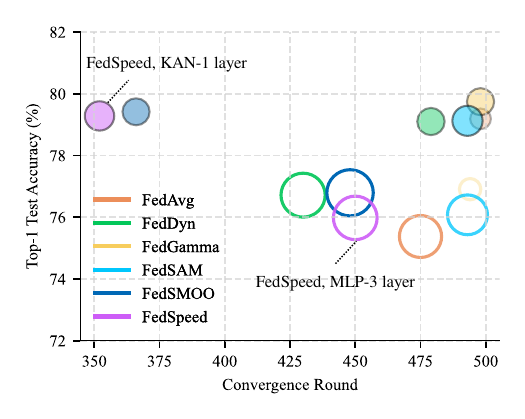}
\caption{Benchmarking results comparing KAN-1 layer (filled circles) and MLP-3 layer (unfilled circles) performance across SOTA FL methods in an IID setting, plotting Convergence Round versus Top-1 Test Accuracy (\%). Circle sizes indicate standard deviation in accuracy.}
\label{fig:intro}
\end{figure}

The Kolmogorov–Arnold Networks (KAN) \cite{liu2024kan1, liu2024kan2}, recently proposed as a potential alternative to Multi-Layer Perceptrons (MLP) \cite{rumelhart1986learning}, can theoretically represent continuous functions without using traditional activation functions like other neural networks.
While recent studies have used KAN for various centralized uses \cite{ji2024comprehensive, somvanshi2024survey}, research on using them in federated settings is still new, mostly limited to the vanilla FedAvg method or using simple datasets \cite{zeydan2024f, zeleke2024federated, sasse2024evaluating}. 
Current research lacks thorough comparisons of KAN over recent FL methods, which prevents a deeper understanding of their practical use in privacy-sensitive medical settings. 
Addressing this gap is crucial to fully understand KAN's potential in complex, spread-out medical data environments. 
Thus, comprehensive testing of KAN across recent FL methods is essential.

To evaluate KAN’s applicability to FL, we compare their performance against MLP-based models across various cutting-edge FL methods, as shown in Fig.~\ref{fig:intro}.
This comparison shows that KAN models, even in their simplest form, perform as well as deeper MLP networks when used with various FL methods in the IID setting.
Additionally, we examine KAN architectures from multiple perspectives, investigating the impact of key hyperparameters—specifically grid size and number of hidden layers—when integrated within FL methods in the Non-IID setting.
These findings will encourage further exploration of KAN's wider use in FL, especially for medical imaging tasks requiring AI interpretability.

The main contributions of this paper are as follows:
\begin{itemize}
    \item We present the first comprehensive benchmark of KAN in FL settings for medical imaging.
    \item We demonstrate that KAN compare favorably with MLP across six state-of-the-art FL methods.
    \item We provide empirical evidence of performance patterns that emerge when varying KAN's key hyperparameters.
    \item In addition, we examine whether increasing KAN's depth through additional hidden layers or expanding its width has greater impact.
\end{itemize}

The remainder of this paper is organized as follows.
In Section \ref{sec:model}, we introduce federated setting and process.
Section \ref{sec:experiment} presents our comprehensive benchmark results across various state-of-the-art FL algorithms.
Finally, Section \ref{sec:conclusion} concludes with remarks.

\section{Problem and Method}\label{sec:model}
\subsection{Federated Setting}
We consider a FL system comprising \(N\) edge devices and a central server as \cite{lee2023fast}. 
Each device \(n \in \{1,2,\dots,N\}\) holds a private dataset \(\mathcal{D}^n\), and due to heterogeneous blood-cell type distributions, the overall data union \(\mathcal{D}=\bigcup_{n=1}^N\mathcal{D}^n\) is inherently imbalanced.  
Consequently, the goal is to learn a global model \(\theta_g\in\Theta\) without aggregating raw data centrally, thereby preserving privacy \cite{lee2022accelerated, li2020federated}.

Accordingly, FL seeks to minimize the empirical risk across all devices:
\[
\min_{\theta_g\in\Theta}
F(\theta_g)
\;\triangleq\;
\frac{1}{N}\sum_{n=1}^N f_n(\theta_g)\,,
\]
where \(F\) denotes the global objective aggregated at the server, and each local objective \(f_n\) on device \(n\) is given by:
\[
f_n(\theta_n)
\;\triangleq\;
\frac{1}{|\mathcal{D}^n|}\sum_{(x_i,y_i)\in\mathcal{D}^n}
\mathcal{L}\bigl(\theta_n; x_i, y_i\bigr)\,,
\]
with \(\mathcal{L}\) representing the per-sample loss (e.g., cross-entropy) for predicting label \(y_i\) from image \(x_i\).

\subsection{Federated Learning Process}

For clarity and concreteness, we describe the FL framework using the FedAvg method.
At each global round \( g_r \in \{1, 2, \dots, G\} \), the central server randomly selects a subset \( \mathcal{S}_{g_r} \subseteq \{1,2,\dots,N\} \) of edge devices for participation and broadcasts the current global model parameters \( \theta_g^{(g_r-1)} \) to these selected devices.

Each participating device \( n \in \mathcal{S}_{g_r} \) then performs local training using its private dataset \( \mathcal{D}^n \) to minimize the local loss function specific to the KAN model. The KAN model computes output \( h(x) \) given input \( x \) via a Kolmogorov–Arnold representation as follows:
\[
h(x) = \sum_{j=1}^{J} c_j \phi_j\left(\sum_{i=1}^{I} w_{ji}\psi_i(x_i) + b_j\right)\,,
\]
where \(I\) and \(J\) denote the dimensions of input and hidden layers, respectively, \(\psi_i\) and \(\phi_j\) represent continuous basis functions of input and hidden layers, and \(w_{ji}\), \(b_j\), and \(c_j\) are learnable weights and biases and output scalars.

The local training on device \( n \) involves minimizing the following local objective through gradient descent \cite{simeone2022machine} for each local epoch \( e \in \{1,2,\dots,E\} \):
\[
\theta_n^{(g_r,e+1)} \leftarrow \theta_n^{(g_r,e)} - \eta \nabla f_n(\theta_n^{(g_r,e)})\,,
\]
where the local loss \( f_n(\theta_n) \) for a given batch \( \mathcal{B}\subseteq \mathcal{D}^n \) is explicitly computed as:
\[
f_n(\theta_n) = \frac{1}{|\mathcal{B}|}\sum_{(x_i,y_i)\in \mathcal{B}}\mathcal{L}\left(h(x_i;\theta_n), y_i\right)\,.
\]

Once local training concludes, each device \( n \) sends its updated local parameters \( \theta_n^{(g_r,E)} \) back to the central server. The server then aggregates these parameters from participating devices using a parameter-wise averaging:
\[
\theta_g^{(g_r)} \leftarrow \frac{1}{|\mathcal{S}_{g_r}|}\sum_{n \in \mathcal{S}_{g_r}} \theta_n^{(g_r,E)}\,,
\]
where each parameter set \(\theta_n\) represents set of parameters \(\{w_{ji}, b_j, c_j\}\) of the KAN model. The aggregated global model \(\theta_g^{(g_r)}\) is then broadcasted back to all devices, initiating the next global round. This federated procedure iteratively continues until convergence criteria for the global model \( \theta_g \) are satisfied. Overall procedure is described in Algorithm \ref{alg}.

\begin{algorithm}
\SetAlgoLined
\DontPrintSemicolon
Initialize global model parameters $\theta_g^{(0)}$\;
\For{global round $g_r = 1, 2, \ldots, G$}{
    Server randomly selects a subset of devices $\mathcal{S}_{g_r} \subseteq \{1, 2, \ldots, N\}$\;
    Server broadcasts $\theta_g^{(g_r-1)}$ to all devices in $\mathcal{S}_{g_r}$\;
    
    \ForPar{device $n \in \mathcal{S}_{g_r}$}{
        Initialize local model: $\theta_n^{(g_r,1)} \leftarrow \theta_g^{(g_r-1)}$\;
        
        \For{local epoch $e = 1, 2, \ldots, E$}{
            Sample batch $\mathcal{B} \subseteq \mathcal{D}^n$\;
            Compute KAN output for each $(x_i, y_i) \in \mathcal{B}$:\;
            $h(x_i;\theta_n^{(g_r,e)}) = \sum_{j=1}^{J} c_j \,\phi_j\!\left(\sum_{i=1}^{I} w_{ji}\,\psi_i(x_i) + b_j\right)$\;
            Compute local loss:\;
            $f_n(\theta_n^{(g_r,e)}) = \frac{1}{|\mathcal{B}|}\!\sum_{(x_i,y_i)\in \mathcal{B}}\!\mathcal{L}\!\left(h(x_i;\theta_n^{(g_r,e)}), y_i\right)$\;
            Update local model:\;
            $\theta_n^{(g_r,e)} \leftarrow \theta_n^{(g_r,e)} - \eta \,\nabla f_n(\theta_n^{(g_r,e)})$\;
        }
        
        Device $n$ sends $\theta_n^{(g_r,E)}$ to the server\;
    }
    
    Server aggregates parameters:\;
    $\theta_g^{(g_r)} \leftarrow \frac{1}{|\mathcal{S}_{g_r}|}\sum_{n \in \mathcal{S}_{g_r}} \theta_n^{(g_r,E)}$\;
}

\Return{$\theta_g^{(G)}$}
\caption{FL with Kolmogorov–Arnold Networks}
\label{alg}
\end{algorithm}

\begin{figure*}[t]
    \centering
    \includegraphics[width=\textwidth]{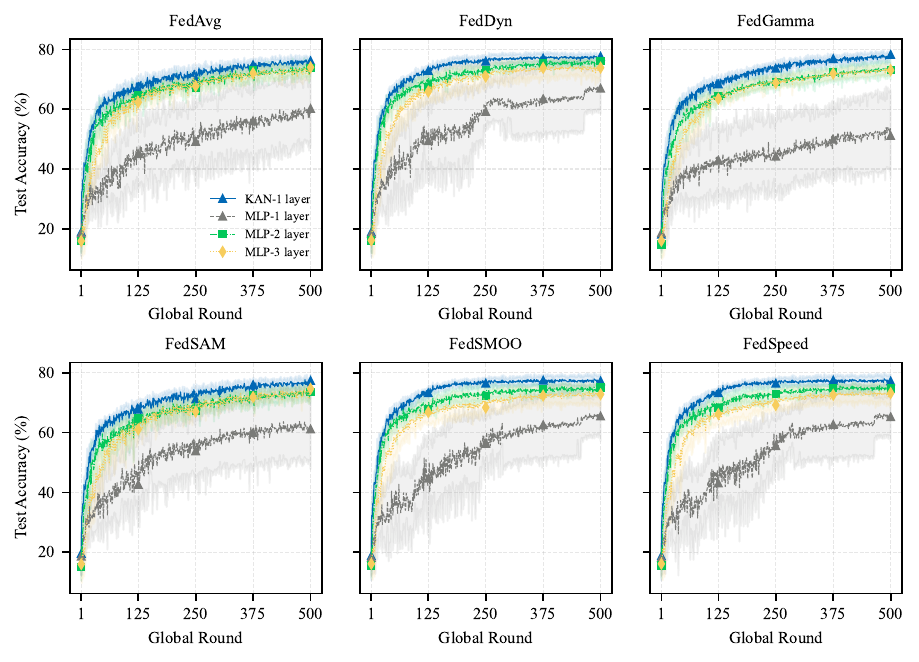}
    \caption{Test accuracy comparison of KAN-1 layer versus MLP architectures (1-3 layers) across six FL algorithms on the blood cell classification dataset with Non-IID distribution ($\alpha=1.0$). Note that shaded regions represent standard deviation across five random seeds.}
    \label{result1}
\end{figure*}

\section{Experiment and Results}\label{sec:experiment}
\subsection{Experiment Setting}
To validate the effectiveness of KAN in federated settings, we conduct experiments on the blood cell classification dataset \cite{acevedo2020dataset} using both KAN and MLP architectures with varying depths.
Note that we employed radial basis function-based FastKAN \cite{li2024kolmogorov} for our experiments to improve computational efficiency.
We examine six state-of-the-art FL algorithms: FedAvg \cite{mcmahan2017communication}, FedDyn \cite{acarfederated}, FedSAM \cite{qu2022generalized}, FedGamma \cite{10269141}, FedSMOO \cite{sun2023dynamic}, and FedSpeed \cite{sunfedspeed} to comprehensively evaluate performance across different federated optimization techniques.
In addition, we simulate a Non-IID data distribution using Dirichlet distribution with $\alpha=1.0$ across $N=100$ edge devices as \cite{li2022federated}, randomly selecting $10\%$ of devices per global round.
We report the mean and standard deviation of test accuracy across five random seeds, and additional hyperparameter settings are available in our open source repository\footnote{https://github.com/yjlee22/fkan}.

\subsection{Impact of KAN on FL Performance}
In this experiment, we aim to explore the potential of KAN as viable replacements for traditional MLP in federated environments.
As shown in Fig. \ref{result1}, KAN consistently outperforms all MLP configurations, suggesting that KAN could potentially replace MLP-based networks in federated environments. 
Furthermore, the single-layer KAN not only achieves higher accuracy than single-layer MLP but also surpasses the performance of deeper MLP architectures while exhibiting more stable convergence across different random seeds. 
This stability is clearly evident from the narrower standard deviation bands in the KAN results compared to MLP. 
Additionally, this advantage remains consistent across all examined FL optimizers, with certain algorithms showing particularly notable synergy with KAN architecture. 
Our findings demonstrate that KAN provide superior performance for Non-IID data in federated settings, making them an effective alternative to MLP for medical imaging applications.

\begin{figure*}[t]
    \centering
    \includegraphics[width=\textwidth]{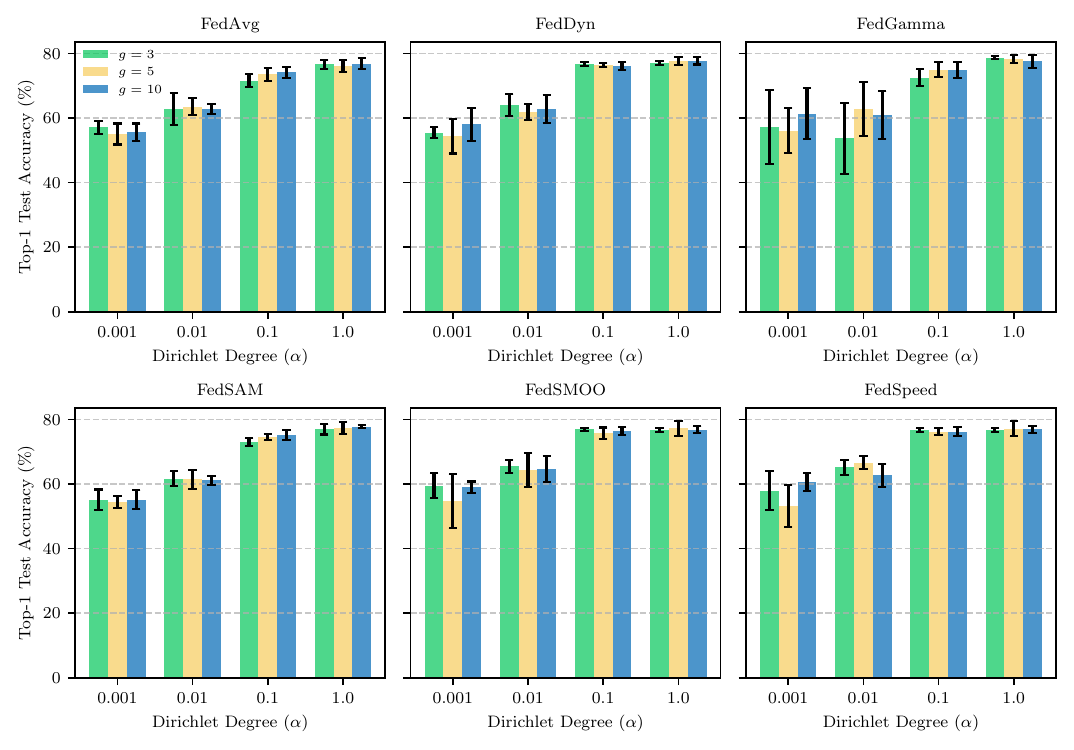}
    \caption{Impact of KAN grid size ($g = 3, 5, 10$) on test accuracy across different Non-IID degrees (Dirichlet parameter $\alpha$ from $0.001$ to $1.0$) for six federated learning algorithms. Lower $\alpha$ values indicate higher data heterogeneity across devices. Note that error bars represent standard deviation across five random seeds.}
    \label{result2}
\end{figure*}

\subsection{Impact of Non-IID Degree}
We investigate how varying the Non-IID degree (Dirichlet parameter $\alpha$) influences the performance of KAN architectures by altering the grid size, as shown in Fig.~\ref{result2}. Results indicate that increasing data heterogeneity (lower $\alpha$) negatively affects test accuracy across all FL methods. 
In detail, larger grid sizes (e.g., $g=10$) generally yield better performance, suggesting grid size is a critical hyperparameter for mitigating Non-IID data challenges. 
Notably, the improvement from larger grids becomes more pronounced as data heterogeneity intensifies, highlighting grid size's importance in highly heterogeneous scenarios. Moreover, specific FL methods, such as FedGamma and FedSpeed, shows notable synergy with larger grid configurations. 
These findings underscore that appropriately selecting the grid size is essential for optimizing KAN performance in diverse federated learning environments.

\begin{table}[t]
\centering
\caption{KAN variants evaluated in the ablation.}
\label{ablation-config}
\renewcommand{\arraystretch}{1.0}
\begin{tabularx}{\columnwidth}{@{}lXlX@{}}
\toprule
\textbf{Config} & \textbf{Hidden-layer} & \textbf{Config} & \textbf{Hidden-layer} \\ 
\midrule
$d_1$ & [5] & $\omega_1$ & [5] \\
$d_3$ & [5, 5, 5] & $\omega_3$ & [125] \\
$d_5$ & [5, 5, 5, 5, 5] & $\omega_5$ & [3125] \\
\bottomrule
\end{tabularx}
\vspace{-0.3cm}
\end{table}

\subsection{Ablation Study}
\begin{figure}[t]
    \centering
    \includegraphics[width=\columnwidth]{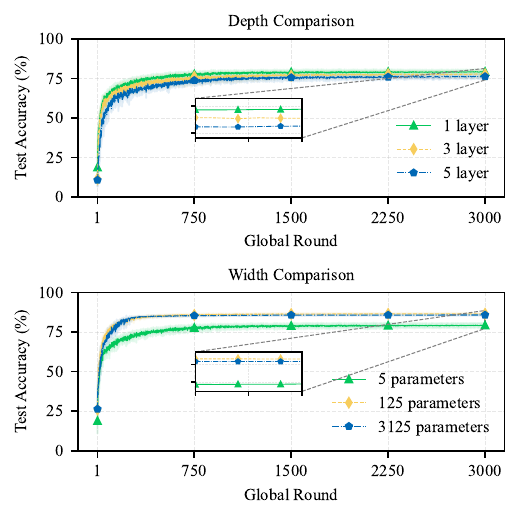}
    \caption{Test accuracy comparison of KAN variants with different depths (top) and widths (bottom) across global rounds.}
    \label{ablation}
\end{figure}

We perform an ablation study using vanilla FedAvg to determine the relative impact of increasing KAN's depth ($d$) versus expanding its width ($\omega$) as Table. \ref{ablation-config}.
As depicted in Fig.~\ref{ablation}, results demonstrate that increasing network depth does not guarantee performance improvements, as shallower KAN architectures (1 layer) actually outperform deeper variants (3 and 5 layers) in our experiments. 
In contrast, width shows significant impact on performance, with the medium configuration (125 parameters) achieving optimal test accuracy compared to both narrow (5 parameters) and wide (3125 parameters) variants. 
The results indicate the importance of carefully tuning width parameters while keeping depth minimal for enhanced KAN performance in FL scenarios. 
Note that depth may have a more significant positive impact than observed in our experiments.

\section{Conclusion}\label{sec:conclusion}
In this work, we present the first comprehensive benchmark of KAN in FL for medical image classification. 
Our results show that KAN can outperform MLP-based network across multiple FL algorithms while maintaining better stability in Non-IID settings. 
Additionally, we find larger grid configurations provide greater benefits as data heterogeneity increases, offering valuable insights for real-world deployment. 
Contrary to conventional design principles, our studies reveal that increasing KAN width yields more benefits than increasing depth for medical imaging tasks. 
Therefore, these findings establish KAN as a promising alternative for privacy-sensitive medical applications, opening new research directions for federated healthcare environments.

\bibliographystyle{IEEEtran}
\bibliography{reference}

\begin{thebibliography}{10}
\providecommand{\url}[1]{#1}
\csname url@samestyle\endcsname
\providecommand{\newblock}{\relax}
\providecommand{\bibinfo}[2]{#2}
\providecommand{\BIBentrySTDinterwordspacing}{\spaceskip=0pt\relax}
\providecommand{\BIBentryALTinterwordstretchfactor}{4}
\providecommand{\BIBentryALTinterwordspacing}{\spaceskip=\fontdimen2\font plus
\BIBentryALTinterwordstretchfactor\fontdimen3\font minus \fontdimen4\font\relax}
\providecommand{\BIBforeignlanguage}[2]{{%
\expandafter\ifx\csname l@#1\endcsname\relax
\typeout{** WARNING: IEEEtran.bst: No hyphenation pattern has been}%
\typeout{** loaded for the language `#1'. Using the pattern for}%
\typeout{** the default language instead.}%
\else
\language=\csname l@#1\endcsname
\fi
#2}}
\providecommand{\BIBdecl}{\relax}
\BIBdecl

\bibitem{bisio2025ai}
I.~Bisio, C.~Fallani, C.~Garibotto, H.~Haleem, F.~Lavagetto, M.~Hamedani, A.~Schenone, A.~Sciarrone, and M.~Zerbino, ``Ai-enabled internet of medical things: Architectural framework and case studies,'' \emph{IEEE Internet Things Mag.}, vol.~8, no.~2, pp. 121--128, Feb. 2025.

\bibitem{bisio2023feet}
I.~Bisio, C.~Garibotto, F.~Lavagetto, and M.~Shahid, ``Feet pressure prediction from lower limbs imu sensors for wearable systems in remote monitoring architectures,'' in \emph{Proc. IEEE GLOBECOM}, Kuala Lumpur, Malaysia, Dec. 2023.

\bibitem{antunes2022federated}
R.~S. Antunes, C.~Andr{\'e}~da Costa, A.~K{\"u}derle, I.~A. Yari, and B.~Eskofier, ``Federated learning for healthcare: Systematic review and architecture proposal,'' \emph{ACM Trans. Intell. Syst. Technol.}, vol.~13, no.~4, pp. 1--23, May 2022.

\bibitem{mcmahan2017communication}
B.~McMahan, E.~Moore, D.~Ramage, S.~Hampson, and B.~A. y~Arcas, ``Communication-efficient learning of deep networks from decentralized data,'' in \emph{Proc. AISTAT}, Fort Lauderdale, United States, Apr. 2017.

\bibitem{li2019convergence}
X.~Li, K.~Huang, W.~Yang, S.~Wang, and Z.~Zhang, ``On the convergence of fedavg on non-iid data,'' in \emph{Proc. ICLR}, Virtual Event, May 2020.

\bibitem{pfitzner2021federated}
B.~Pfitzner, N.~Steckhan, and B.~Arnrich, ``Federated learning in a medical context: a systematic literature review,'' \emph{ACM Trans. Internet Technol.}, vol.~21, no.~2, pp. 1--31, June 2021.

\bibitem{lee2024security}
Y.~Lee, S.~Park, and J.~Kang, ``Security-preserving federated learning via byzantine-sensitive triplet distance,'' in \emph{Proc. IEEE ISBI}, Athens, Greece, June 2024.

\bibitem{kairouz2021advances}
P.~Kairouz and H.~McMahan, \emph{Advances and Open Problems in Federated Learning}, ser. Found. Trends Mach. Learn.\hskip 1em plus 0.5em minus 0.4em\relax Now Publishers, June 2021, vol.~14.

\bibitem{lee2025revisit}
Y.~Lee, J.~Gong, S.~Choi, and J.~Kang, ``Revisit the stability of vanilla federated learning under diverse conditions,'' \emph{arXiv preprint arXiv:2502.19849}, 2025.

\bibitem{liu2024kan1}
Z.~Liu, Y.~Wang, S.~Vaidya, F.~Ruehle, J.~Halverson, M.~Solja{\v{c}}i{\'c}, T.~Y. Hou, and M.~Tegmark, ``Kan: Kolmogorov-arnold networks,'' in \emph{Proc. ICLR}, Singapore, Apr. 2025.

\bibitem{liu2024kan2}
Z.~Liu, P.~Ma, Y.~Wang, W.~Matusik, and M.~Tegmark, ``Kan 2.0: Kolmogorov-arnold networks meet science,'' \emph{arXiv preprint arXiv:2408.10205}, 2024.

\bibitem{rumelhart1986learning}
D.~E. Rumelhart, G.~E. Hinton, and R.~J. Williams, ``Learning representations by back-propagating errors,'' \emph{Nature}, vol. 323, no. 6088, pp. 533--536, Oct. 1986.

\bibitem{ji2024comprehensive}
T.~Ji, Y.~Hou, and D.~Zhang, ``A comprehensive survey on kolmogorov arnold networks (kan),'' \emph{arXiv preprint arXiv:2407.11075}, 2024.

\bibitem{somvanshi2024survey}
S.~Somvanshi, S.~A. Javed, M.~M. Islam, D.~Pandit, and S.~Das, ``A survey on kolmogorov-arnold network,'' \emph{arXiv preprint arXiv:2411.06078}, 2024.

\bibitem{zeydan2024f}
E.~Zeydan, C.~J. Vaca-Rubio, L.~Blanco, R.~Pereira, M.~Caus, and A.~Aydeger, ``F-kans: Federated kolmogorov-arnold networks,'' in \emph{Proc. IEEE CCNC DAINET WorkShop}, Las Vegas, United States, Jan. 2024.

\bibitem{zeleke2024federated}
S.~N. Zeleke and M.~Bochicchio, ``Federated kolmogorov-arnold networks for health data analysis: A study using ecg signal,'' in \emph{Proc. IEEE BigData}, Washington, United States, Dec. 2024.

\bibitem{sasse2024evaluating}
A.~M. Sasse and C.~M. de~Farias, ``Evaluating federated kolmogorov-arnold networks on non-iid data,'' \emph{arXiv preprint arXiv:2410.08961}, 2024.

\bibitem{lee2023fast}
Y.~Lee, S.~Park, and J.~Kang, ``Fast-convergent federated learning via cyclic aggregation,'' in \emph{Proc. IEEE ICIP}, Kuala Lumpur, Malaysia, Oct. 2023.

\bibitem{lee2022accelerated}
Y.~Lee, S.~Park, J.-H. Ahn, and J.~Kang, ``Accelerated federated learning via greedy aggregation,'' \emph{IEEE Commun. Lett.}, vol.~26, no.~12, pp. 2919--2923, Dec. 2022.

\bibitem{li2020federated}
T.~Li, A.~K. Sahu, A.~Talwalkar, and V.~Smith, ``Federated learning: Challenges, methods, and future directions,'' \emph{IEEE Signal Process. Mag.}, vol.~37, no.~3, pp. 50--60, May 2020.

\bibitem{simeone2022machine}
O.~Simeone, \emph{Machine learning for engineers}.\hskip 1em plus 0.5em minus 0.4em\relax Cambridge University Press, 2022.

\bibitem{acevedo2020dataset}
A.~Acevedo, A.~Merino, S.~Alf{\'e}rez, {\'A}.~Molina, L.~Bold{\'u}, and J.~Rodellar, ``A dataset of microscopic peripheral blood cell images for development of automatic recognition systems,'' \emph{Data Br.}, vol.~30, June 2020.

\bibitem{li2024kolmogorov}
Z.~Li, ``Kolmogorov-arnold networks are radial basis function networks,'' \emph{arXiv preprint arXiv:2405.06721}, 2024.

\bibitem{acarfederated}
D.~A.~E. Acar, Y.~Zhao, R.~Matas, M.~Mattina, P.~Whatmough, and V.~Saligrama, ``Federated learning based on dynamic regularization,'' in \emph{Proc. ICLR}, Vienna, Austria, May 2021.

\bibitem{qu2022generalized}
Z.~Qu, X.~Li, R.~Duan, Y.~Liu, B.~Tang, and Z.~Lu, ``Generalized federated learning via sharpness aware minimization,'' in \emph{Proc. ICML}, Baltimore, United States, Jul. 2022.

\bibitem{10269141}
R.~Dai, X.~Yang, Y.~Sun, L.~Shen, X.~Tian, M.~Wang, and Y.~Zhang, ``Fedgamma: Federated learning with global sharpness-aware minimization,'' \emph{IEEE Trans. Neural Netw. Learn. Syst.}, vol.~35, no.~12, pp. 17\,479--17\,492, Dec. 2024.

\bibitem{sun2023dynamic}
Y.~Sun, L.~Shen, S.~Chen, L.~Ding, and D.~Tao, ``Dynamic regularized sharpness aware minimization in federated learning: Approaching global consistency and smooth landscape,'' in \emph{Proc. ICML}, Hawaii, United States, June 2023.

\bibitem{sunfedspeed}
Y.~Sun, L.~Shen, T.~Huang, L.~Ding, and D.~Tao, ``Fedspeed: Larger local interval, less communication round, and higher generalization accuracy,'' in \emph{Proc. ICLR}, Kigali, Rwanda, May 2023.

\bibitem{li2022federated}
Q.~Li, Y.~Diao, Q.~Chen, and B.~He, ``Federated learning on non-iid data silos: An experimental study,'' in \emph{Proc. IEEE ICDE}, Kuala Lumpur, Malaysia, May 2022.

\end{thebibliography}

\end{document}